\documentclass[letterpaper, 10 pt, conference]{ieeeconf}  

\IEEEoverridecommandlockouts                              

\usepackage{graphics} 
\usepackage{epsfig} 
\usepackage{mathptmx} 
\usepackage{times} 
\usepackage{amsmath} 
\usepackage{amssymb}  
\usepackage{algorithm}
\usepackage{algpseudocode}
\usepackage{booktabs}

\title{\LARGE \bf
Stimulating Imagination: Towards General-purpose \\ ``Something Something Placement''
}

\author{
  Jianyang Wu$^{*}$$^\dag$
  \quad
  Jie Gu$^{*}$$^\ddag$
  \quad
  Xiaokang Ma$^{*}$
  \quad
  Fangzhou Qiu
  \quad
  Chu Tang
  \quad
  Jingmin Chen
  \thanks{All authors are with the Rightly Robotics, Hangzhou, China. (e-mail: jgu@rightly.ai; xma@rightly.ai; jingmin.chen@rightly.ai).}
  \thanks{$^{*}$Equal contribution. $^\dag$Work completed during the internship at Rightly Robotics. $^\ddag$Corresponding Author.}
}

\begin{document}

\maketitle
\thispagestyle{empty}
\pagestyle{empty}

\begin{abstract}
  General-purpose object placement is a fundamental capability of an intelligent generalist robot: being capable of rearranging objects following precise human instructions even in novel environments. This work is dedicated to achieving general-purpose object placement with ``something something'' instructions. Specifically, we break the entire process down into three parts, including object localization, goal imagination and robot control, and propose a method named SPORT. SPORT leverages a pre-trained large vision model for broad semantic reasoning about objects, and learns a diffusion-based pose es-timator to ensure physically-realistic results in 3D space. Only object types (movable or reference) are communicated between these two parts, which brings two benefits. One is that we can fully leverage the powerful ability of open-set object recognition and localization since no specific fine-tuning is needed for the robotic scenario. Moreover, the diffusion-based estimator only need to ``imagine" the object poses after the placement, while no necessity for their semantic information. Thus the training burden is greatly reduced and no massive training is required. The training data for the goal pose estimation is collected in simulation and annotated by using GPT-4. Experimental results demonstrate the effectiveness of our approach. SPORT can not only generate promising 3D goal poses for unseen simulated objects, but also be seamlessly applied to real-world settings.
\end{abstract}

\section{INTRODUCTION}

General-purpose object placement is a fundamental capability of an intelligent generalist robot. This work focuses on the ``something something placement'' task: place some object relative to another one. Much like humans, the robot must be able to recognize target objects (even though it has never encountered before), and then rearranging the objects following human instructions. For example, if an instruction ``put the spicy potato chips on the plate'' is given, the ability of semantic understanding is required, \emph{i.e.}, reasoning about ``spicy potato chips'' even this phrase may be outside of the training distribution. Furthermore, the rearrangement should be physically-realistic by fully considering the physical structures, geometries and constraints. 

The great progresses of generative models provide an insight of solving this challenging problem. Some of them introduce powerful models pre-trained on vision \cite{RT2,susie,genaug,ROSIE}, for initializing robotic policies or enhancing semantic understanding. Despite benefiting from the large-scale pre-training, it remains doubts about the generalization ability, since the amount of robotic manipulation fine-tuning data are far less than that encountered in a person's experience (also less than pre-training). Another bottleneck is that these approaches do not look specifically at 3D spatial understanding. They assume that the underlying states of world can be characterized by 2D images from certain angles.

Some other works directly learn the object rearrangement skill in the 3D space \cite{StructFormer,StructDiffusion}. Specifically, the model takes point clouds as inputs and learns to directly estimate goal poses of rearranged objects. The training data are collected with physics simulators to ensure physically-valid results. However, given the fact that obtaining simulation data is expensive and time-consuming, the sizes of existing datasets are limited and scaling robot learning is difficult. Accordingly, these models lack the broad semantic understanding and reasoning ability for object localization, and may also fail to follow precise low-level instructions.

\begin{figure}
  \centering
  \includegraphics[width=0.45\textwidth]{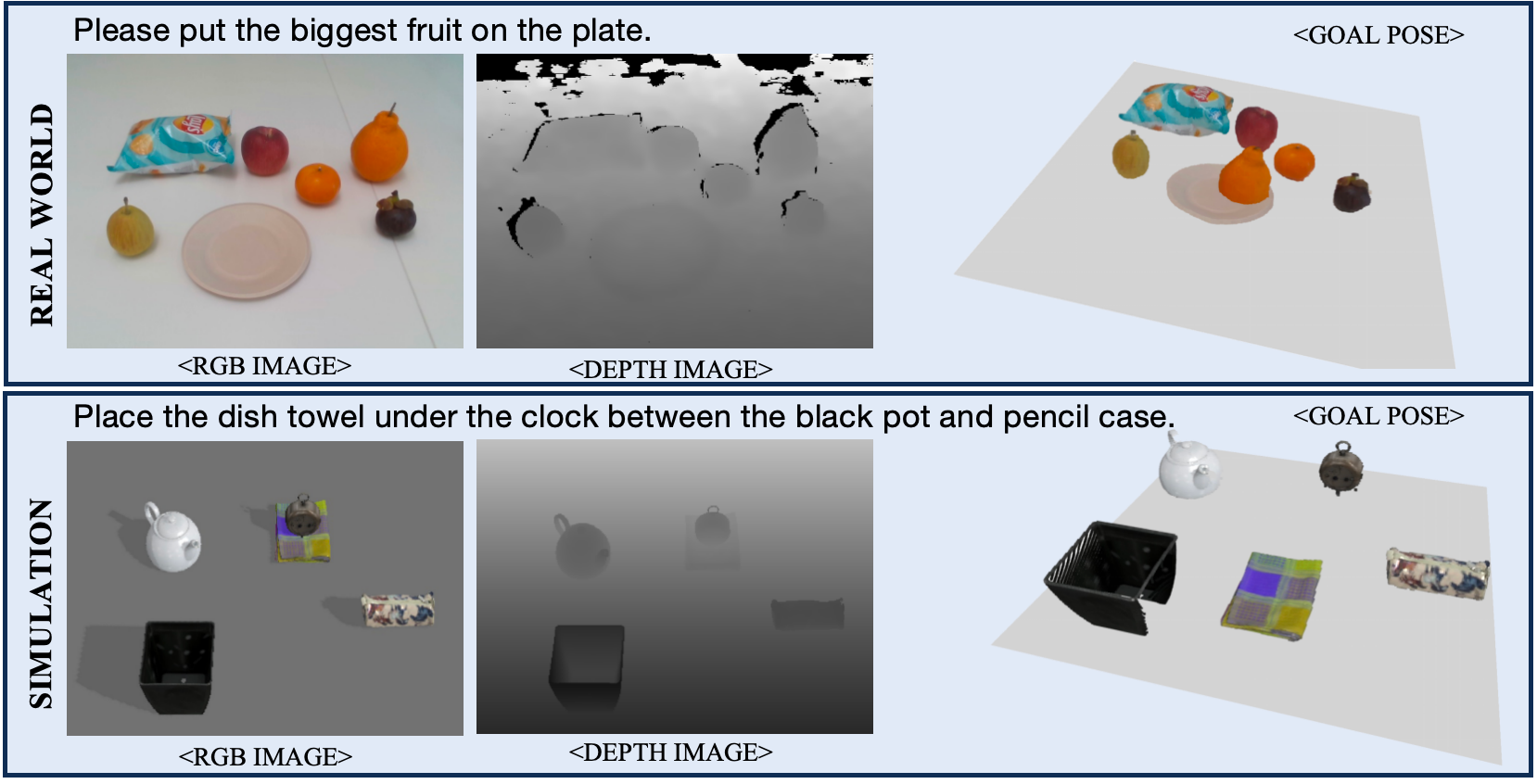}
  \caption{SPORT can ``imagine'' the 3D poses of the rearranged objects given a language instruction, being semantic-aware and physically-realistic.}
  \label{fig:intro}
\end{figure}

This paper presents a framework that leverages pre-trained large vision model and 3D spatial reasoning model to enable general-purpose object rearrangement with ``something something'' instructions. The key insight is that the process can be decoupled into three parts, namely object recognition and localization, goal imagination and robot control. We concentrate on the first two parts (robot control can be done via separated learned control policies \cite{susie} or Model Predictive Control). Without loss of generality, we use ``put the spicy potato chips on the plate'' as an example. The movable object (spicy potato chips) and the reference object (plate) are first recognized and localized by using an LLM-enhanced open-set segmentation algorithm \cite{LISA,SAM}, fully utilizing the powerful seasoning and semantic understanding ability derived from large-scale pre-training. The corresponding partial-view point clouds of these objects are obtained based on segments and RGB-D inputs. A natural language conditioned diffusion \cite{DDPM} model is then introduced to ``imagine'' the goal poses in 3D space that satisfy the rearrangement instruction and physical constraints. Note that the diffusion model can predict the object poses by only accessing their types, \emph{i.e.}, to be moved or reference, while without semantic information. This ease the burden of training, allowing us to learn well-generalized models with relatively little data.

In particular, the contributions of this work are as follows.
\begin{itemize}
    \item We demonstrate the potential to effectively address the ``place something relative to something'' task. The key is the ability of reasoning and recognizing various objects, as well as understanding 3D spatial relationships for a physically-realistic placement. Corresponding modules can be decoupled.
    \item We present our approach, which leverages a powerful vision model pre-trained on large-scale data for general-purpose object localization, and develops a diffusion-based model for physically-realistic pose estimation in 3D space. The results from both simulation and real-world experiments demonstrate the effectiveness of our approach, even on novel objects in novel environments.
    \item One key challenge is the lack of 3D object rearrangement data containing precise and low-level instructions. We establish a GPT-assisted pipeline from a 3D perspective to generate high-quality data. The data is generated in simulation to ensure physical realism.
\end{itemize}

\section{Related work}

\textbf{2D pose estimation.} 
Traditionally, the object rearrangement task is divided into object recognition and pose estimation tasks. The advancement of vision transformers has greatly improved object recognition performance, making robots more semantically aware. Neural networks are also used to generate object poses from 2D images and relational predicates as inputs \cite{meesLearningObjectPlacements2020,venkateshSpatialReasoningNatural2021,kartmannRepresentingSpatialObject2020,wuTransportersVisualForesight2022,yuanSORNetSpatialObjectCentric2022}. However, two challenges remain: first, the set of relational predicates is limited; second, these methods struggle to handle collisions in 2D settings.

\textbf{3D scene generation.}
Vision generation models have endowed AI with visual imagination capabilities. Some research has attempted to apply diffusion models to arrangement tasks, thereby generalizing placement instructions. For instance, DALL-E-BOT \cite{kapelyukhDALLEBotIntroducingWebScale2023} generates a goal image from text and matches it with an observed image to determine new object positions. However, it uses diffusion models directly, resulting in generated images that often differ from observed ones. Dream2Real \cite{kapelyukhDream2RealZeroShot3D2023} employs a sampling strategy to sample candidate object positions in 3D space and uses Vision Language Models to score each candidate. Working in 3D space can make the generated scenes more physically realistic, avoiding collisions or placing objects in mid-air. StructFormer \cite{StructFormer} and StructDiffusion \cite{StructDiffusion} take a more direct approach, using transformers and diffusion models to edit observed 3D point clouds, enabling the execution of abstract instructions such as "set the dining table." However, they exhibit weak referential capabilities for objects.

\textbf{Imitation Learning.}
Another research direction involves generating robotic actions directly. For example, Transporter \cite{zengTransporterNetworksRearranging2022} uses ResNet to generate robotic actions instead of object poses, while CLIP-PORT \cite{shridharCLIPortWhatWhere2021} combines CLIP \cite{clip} and Transporter to enhance object recognition capabilities. With the development of Multimodal Large Models, some works address robotic manipulation problems more end-to-end. Examples include Diffusion Policy \cite{chiDiffusionPolicyVisuomotor2023}, VIMA \cite{jiangVIMAGeneralRobot2023}, and RT2 \cite{RT2}, which attempt to solve general robotic manipulation with language prompts and visual observations to generate visuomotor actions directly. However, these imitation approaches usually require large amounts of robotic tele-operation or simulation data to achieve generalization capabilities.

\begin{figure*}
  \centering
  \includegraphics[width=0.895\textwidth]{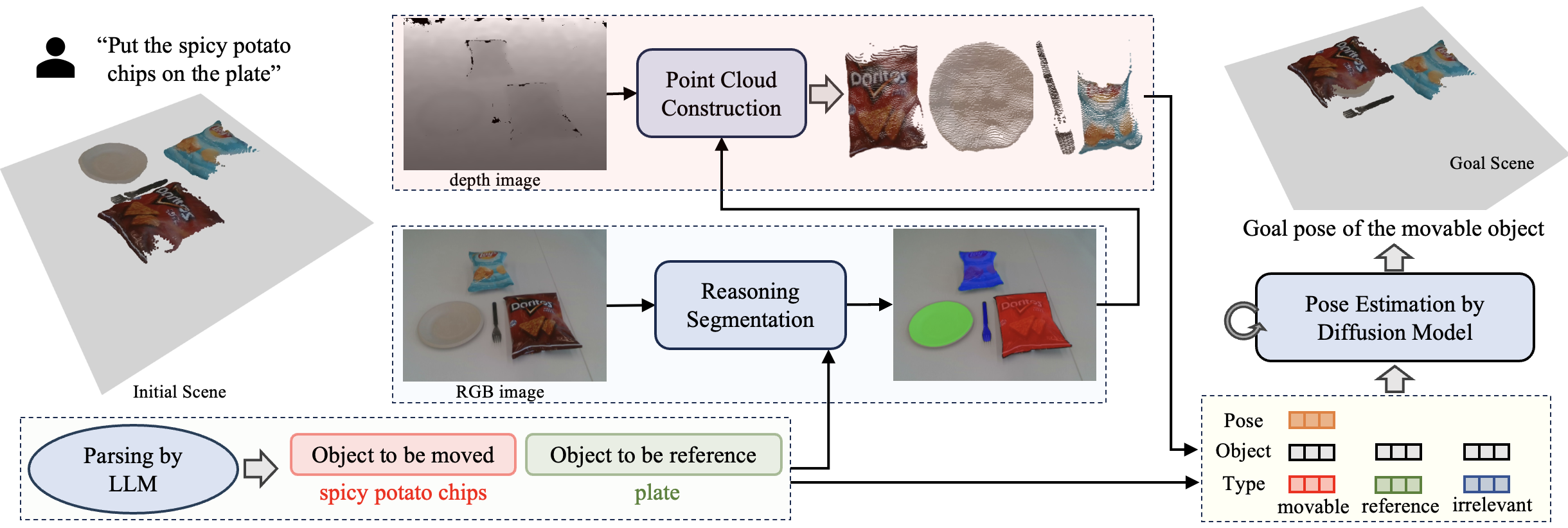}
  \caption{The pipeline of SPORT. The inputs are the RGB-D images of the initial scene and the user command. The movable and reference objects are first extracted (``spicy potato chips'' can be correctly obtained even with the distraction of chips in other flavors). Then a transformer-based diffusion model estimates the goal pose of the ``spicy potato chips'' in 3D space, satisfying the command and being physically-realistic.
  }
  \label{Fig:main}
\end{figure*}

\section{Semantic-aware and physically-realistic object rearrangement}

We introduce SPORT, a \underline{S}emantic-aware and \underline{P}hysically-realistic \underline{O}bject \underline{R}errangemen\underline{T} method. The pipeline of SPORT is illustrated in Fig. \ref{Fig:main}.

\subsection{Preliminaries and Problem Formulation}
Given a single view of a scene captured by RGB-D sensors, we wish for the robot to rearrange this scene to satisfy the natural language instruction. Let $\textbf{I}_{rgb}$ and $\textbf{I}_d$ be the captured RGB and depth images, respectively. Denote the $N$ objects in the scene as $\textbf{O}=\{O_1,O_2,\ldots,O_N\}$, and the language instruction as $\textbf{W}$. Typically, the robot would be required to treat some object as the reference, and then move another one to a certain position relative to the reference. 

Our key insight is that decoupling the components of rearrangement helps improve generalization. First, being capable of reasoning about novel objects and scenarios is necessary, even they may not be present in the robotic training data. A large language model (LLM) enhanced segment anything model (trained on Internet-scale data) is introduced to enable this. Specifically, the object types $\textbf{T}$ $=\{T_1,T_2,\ldots,T_N\}$ are inferred from $\textbf{W}$. Without loss of generality, we use subscripts $r$ and $m$ to denote the reference and movable objects, \emph{i.e.}, $O_r$ and $O_m$ respectively. The remaining objects, namely $\textbf{O} \setminus \{O_r,O_m\}$, are irrelevant ones for the given command. The vision segmentation model takes $\textbf{I}_{rgb}$ and object descriptions as inputs and output segmentation masks of objects, denoted by $\textbf{M}=\{M_1,M_2,\ldots,M_N\}$. According to the $\textbf{M}$ and $\textbf{I}_d$, the partial-view point clouds of objects can be derived, denoted by $\textbf{P}=\{P_1,P_2,\ldots,P_N\}$.

Then the robot should act like a human that can ``imagine'' the 3D goal poses of objects after the movement. A diffusion-based model is developed to achieve this, which takes $\textbf{P}$, $\textbf{W}$ and $\textbf{T}$ as inputs and estimate the 3D pose $\mathbf{x}^m$ (only the object to be moved requires the pose estimation). One should note that the diffusion-based model only accesses to the object types (which object requires to be moved and which one is the reference) while without knowing what they are. ``Moving $a$ to the left of $b$'' and ``moving $c$ to the left of $d$'' are nearly equivalent to the pose estimation in our setting, because the final relative positions of objects are the same in these two commands. Though without semantic information, the physical validity of the rearrangement can still be ensured by understanding point cloud data. Benefiting from this, a well-generalized goal pose estimator can be trained without massive data.
\footnote{After obtaining the goal poses, the robot control can be done via mature methods like MPC. Since this is not the focus of our work, related statements are not included.}

\subsection{SPORT for Something Something Placement}

\textbf{Instruction parsing} An LLM (\emph{e.g.}, GPT \cite{GPT4,GPT3} or LLaMA \cite{llama2}) is utilized to understand and parse the natural-language command $\textbf{W}$. Still use ``put the spicy potato chips on the plate'' as an instance: ``Spicy potato chips'' is the object to be moved and ``plate'' is the reference.

\textbf{Object reasoning and segmentation} An open-set segmentation model is then needed. We employ LISA \cite{LISA} in this work though other similar models could also be used. LISA combines LLaVA \cite{llava} (multi-modal LLM ) with SAM \cite{SAM} (segmentation decoder), showing powerful capacity of complex semantic reasoning that requires world knowledge. For example, given an image containing two bags of potato chips, it can segment the spicy one according to the common knowledge ``spicy snacks are usually packaged in red''. 

We do not fine-tune the vision model on robotic data. The reason is that the scale of robotic data is smaller than that of web data for pre-training the large vision model \cite{RT2,ROSIE}. We can fully leverage the existing high-capacity of complex reasoning and semantic understanding, while fine-tuning may hurt the generalization. 

\textbf{Pose estimation} We parameterize the 6-DoF pose as $(s,R)$ $\in SE(3)$. The goal pose estimator is based on a diffusion model. Note that unlike recent AIGC approaches \cite{stable_diffusion,controlnet}, the diffusion model here is to generate the 6-DoF poses of the objects in the real-world, rather than the scene image directly. It consists of several modules, namely a general-purpose text encoder, learnable type embeddings, a point cloud encoder and a vanilla transformer \cite{transformer} as the backbone. We only use certain object (the moving one) to train the model, since the other objects remain unchanged after the rearrangement. The underlying idea behind this is similar to inpainting.

\textit{Text encoder}. We deploy BERT \cite{bert} as the text encoder along with its tokenizer. Unlike several previous works that are limited to the tokens in their customized vocabulary \cite{StructDiffusion}, BERT can help the model to understand general-purpose instructions in natural language form.

\textit{Type embedding}. A set of learnable embeddings is introduced to indicate the token types in the placement, \emph{i.e.}, $\textbf{T}$. Four types are considered: the texts, the objects to be moved, the reference and irrelevant objects. With such embeddings, the model can differentiate whether the 6-DoF poses of corresponding objects need to be changed.

\textit{Object encoder}. There are two types of object representations (concatenated). One encodes geometric and spatial information by Point Cloud Transformer (PCT) \cite{PCT}, given segmented partial-view point clouds of objects \textbf{P}. The other one encodes pose information at the last time-step of diffusion model: using a multi-layer perceptron (MLP) to encode $(s,R)$. Apart from movable objects, the poses of other objects remain consistent with their initial pose. 

\textit{Diffusion}. A language-conditioned diffusion model is used to estimate the goal poses of objects. At each time-step, six types of embeddings (specifically the text, type, object, position, time and an extra token containing camera viewpoint information) are combined and fed to the backbone. The model then predict the poses at the current time-step, specifically the $s \in R^3$ and two vectors $a,b \in R^3$ to construct the rotation matrix $R \in SO(3)$. The position and time embeddings follow standard design, indicating the token positions in sequences and the time-step in diffusion, respectively. 

All objects are included in model inputs, but only the one to be moved gets involved in iterative pose estimation. It is because the diffusion model needs to access all object information to achieve relative positional movement and avoid collisions, while only the pose of the movable object would change. Accordingly, we only add noise to the movable object pose during model training. Within the DDPM framework \cite{DDPM}, the training objective can be formulated as 
\begin{equation}
\underset{\theta}{\mathrm{arg \, min}} \sum_{i=1}^{N} \, \mathbb{I}_{i=m} \, \mathbb{E}_{\mathbf{\epsilon}, t} \left[ \left\| \mathbf{\epsilon} - \mathbf{\epsilon}_{\theta}(\mathbf{x}^i_t, t, \textbf{W}, \textbf{T}, \textbf{P}) \right\|_1 \right],
\end{equation}
where \( \mathbf{\epsilon} \) is sampled from a standard normal distribution, $\mathbf{x}^i_t$ is the pose estimation of $i$-th object in $\textbf{O}$ at $t$-timestep, and $\mathbb{I}_{i=m}$ is an indicator checking whether $i$-th object is the one to be moved. The generation process of goal pose is shown in Algorithm~\ref{alg:sample_pe}. It directly estimates the 6-DoF poses of the movable object in the 3D space.

\begin{algorithm}
\caption{3D Goal Pose Generation}
\label{alg:sample_pe}
\centering
\begin{algorithmic}[1]
    \State $\mathbf{x}_T^m \sim N(\mathbf{0},\mathbf{I})$
    \For{$t$ =\ $T,\ \ldots,\ 1$}
        \State $\mathbf{z} \sim N(\mathbf{0},\mathbf{1})$ if $t$\ \textgreater\ 1,\  else $\mathbf{z}=\mathbf{0}$
        \State $\mathbf{x}_{t-1}^m=\frac{1}{\sqrt{\alpha_t}}(\mathbf{x}_t^m-\frac{1-\alpha_t}{\sqrt{1-\alpha_t}} \mathbf{\epsilon}_{\theta}(\mathbf{x}_t^m, t, \textbf{W}, \textbf{T}, \textbf{P}))+\sigma_t\mathbf{z}$
    \EndFor
    \State return $\mathbf{x}_0^m$
\end{algorithmic}
\end{algorithm}

\subsection{GPT-assisted Data Generation}
\label{sec:data_gen}
The public available data for ``place something relative to something'' is limited. In this work, an automatic pipeline is proposed for collecting high-quality placement-instruction pairs. Each instance comprises an initial scene, a goal scene after the rearrangement and a corresponding instruction. A total of 40K stable and collision-free instances are generated in the PyBullet physics simulator \cite{PyBullet}, rendered by OpenGL \cite{opengl}. The simulated objects are randomly selected from the popular ShapeNetSem \cite{shapenet2015,savva2015semgeo} dataset. We collect various 581 objects from 30 categories to ensure diversity.

The pipeline consists of three steps: (1) pre-processing metadata to obtain well-constructed and realistic simulated objects; (2) randomly selecting the reference, movable and irrelevant objects, namely $O_r$, $O_m$ and $\textbf{O} \setminus \{O_r,O_m\}$, loading them in PyBullet to obtain the initial and goal scenes (also filtering out physically-unrealistic ones); (3) using GPT-4 to generate language instructions corresponding to the transition from the initial to the goal scenes. The difficulty lies in the time-consuming and cumbersome data collection process, as well as the limited capability of precise (fine-grained) spatial understanding in existing models, even GPT-4.

\textit{Pre-processing}. The object models in ShapeNetSem may have unrealistic sizes and the centroids of them may not be aligned with the origin of the point-cloud coordinate system. Thus scaling and translation are required. The scaling can be done with GPT-4, \emph{e.g.}, asking GPT-4 about the typical size of a cellphone and accordingly scaling the object model. 

\textit{Scene generation}.
We build multiple scenarios according to the relative spatial relationship between the movable and reference objects, such as left, right, between, etc. The object set $\textbf{O}$ and the scenario are randomly selected to generate scenes. For the initial scene, we place all objects with random positions in PyBullet, recording the poses once they are stabilized. For the goal scene, we replace the reference and irrelevant objects with the recorded poses and then load $O_m$. Its pose is randomly sampled within a region. For example, in the coordinate system with $O_r$ as the origin, ``left'' refers to the region $\{ (x, y) \in \mathbb{R}^2 \,|\, x/\sqrt{x^2+y^2} < -\delta, |y|/\sqrt{x^2+y^2} < \delta \}$ ($\delta$ is a hyper-parameter). The physical validity refers to stability and collision. They are measured by angular and linear velocities in physics engine, and checking whether the positions of $\textbf{O} \setminus \{O_m\}$ has any displacement, respectively.

\textit{Instruction generation}.
Inspired by previous works \cite{llava}, we use GPT4 to generate natural language instructions, given the spatial coordinates and object information (\emph{e.g.}, RGB value and size) of scenes. We observe that without such rich information, GPT may probably fail to understand the spatial transition or determine whether the placement is reasonable.

Another interesting observation is that we have tried an end-to-end instruction generation approach: directly prompting GPT4 the rendered RGB images of the initial and final scenes. However, the generated language instructions are of low accuracy. We attribute this to two main reasons: (1) the absence of lifelike qualities in simulated objects impedes GPT4's recognition ability; (2) deducing fine-grained 3D spatial relationships from only RGB images is challenging for existing LLMs \cite{CounterCurate} (even the powerful GPT4).

\section{Experiments}

The goal of the experiments is to evaluate the efficacy of SPORT on the task of place something relative to something, especially in the ability of generalization, 3D spatial reasoning and precise instruction following. To this end, we need to answer the following questions:

\textit{1. Does SPORT excel at the task, even in a new environment, given a precise instruction, given unseen objects with various attributes, requiring physically-realistic results?}

\textit{2. Can SPORT trained with simulation data seamlessly transfer to real-world scenarios, even in zero-shot and requiring more complex reasoning?}

The evaluation are performed in both simulation and real-world environments.

\subsection{Simulation Experiments}

\textbf{Setup} A cross-dataset evaluation is conducted to fully validate the generalization ability of SPORT. \textit{Different from} the training data (ShapeNetSem), the simulated objects for testing are collected from Google Scanned Objects \cite{downs2022google}. A total of 77 object models from 37 novel categories are randomly selected. We use PyBullet \cite{PyBullet} as the physics simulator and OpenGL \cite{opengl} as the appearance render. 4K testing instances (scene and instruction) are generated following the pipeline in subsection \ref{sec:data_gen}, within which the involved objects are randomly sampled.

As in previous works \cite{StructDiffusion}, evaluation metric is the success rate. There are three aspects to consider: given a command, the model should recognize the involved objects, place them to correct positions, and the rearrangement is physically realistic. We systematically measure whether the placed positions of objects satisfy the command, with similar rules for assessing spatial relationships in subsection \ref{sec:data_gen}. To assess physical validity, we continuously place objects in simulation based on the estimated poses, checking the collision and stability. A placement is considered as correct \emph{ONLY} when all these aspects are satisfactory.

A single diffusion model is trained for all scenarios of spatial relationships. We strive to ensure that the data amount of each scenario is balanced. Adam optimizer is used with a learning rate of 1e-4. The batch size is set to be 256. The training is performed for 200 epochs, which takes 4 hours on a single A100 GPU. During the inference phase, we use 200 steps for the denoising process of the diffusion model.

\begin{table}
  \caption{\textbf{(Performance in simulation)} All testing objects are unseen to the model. ``Pose Accuracy'' refers to the accuracy of whether the final object poses satisfy the instruction. ``Physical Realism'' refers to the accuracy of whether the placement is physically realistic. ``Overall Success'' is achieved only when the above two requirements are met.}
  \label{table:1}
  \centering
  \begin{tabular}{l|c|c|c}
  \toprule
    \multicolumn{1}{c}{} & \multicolumn{1}{|c}{Pose Accuracy}  & \multicolumn{1}{|c}{Physical Realism} & \multicolumn{1}{|c}{Overall Success} \\
    \midrule
    SPORT & 59.64 & 70.48 & 46.19 \\
    SPORT* & 87.80 & 76.40 & 69.49     \\
    \midrule
    Poses-train & 83.46 & 77.68 & 65.59  \\
    BERT-train  & 36.38 & 75.04 & 27.65  \\
    \midrule
    50\% data & 80.48 & 72.16 & 62.42  \\
    25\% data & 76.17 & 71.59 & 58.12  \\
    10\% data & 64.89 & 63.17 & 46.73 \\
    \bottomrule
  \end{tabular}
\end{table}

\textbf{Quantitative evaluation}

\begin{figure}
  \centering
  \includegraphics[width=0.45\textwidth]{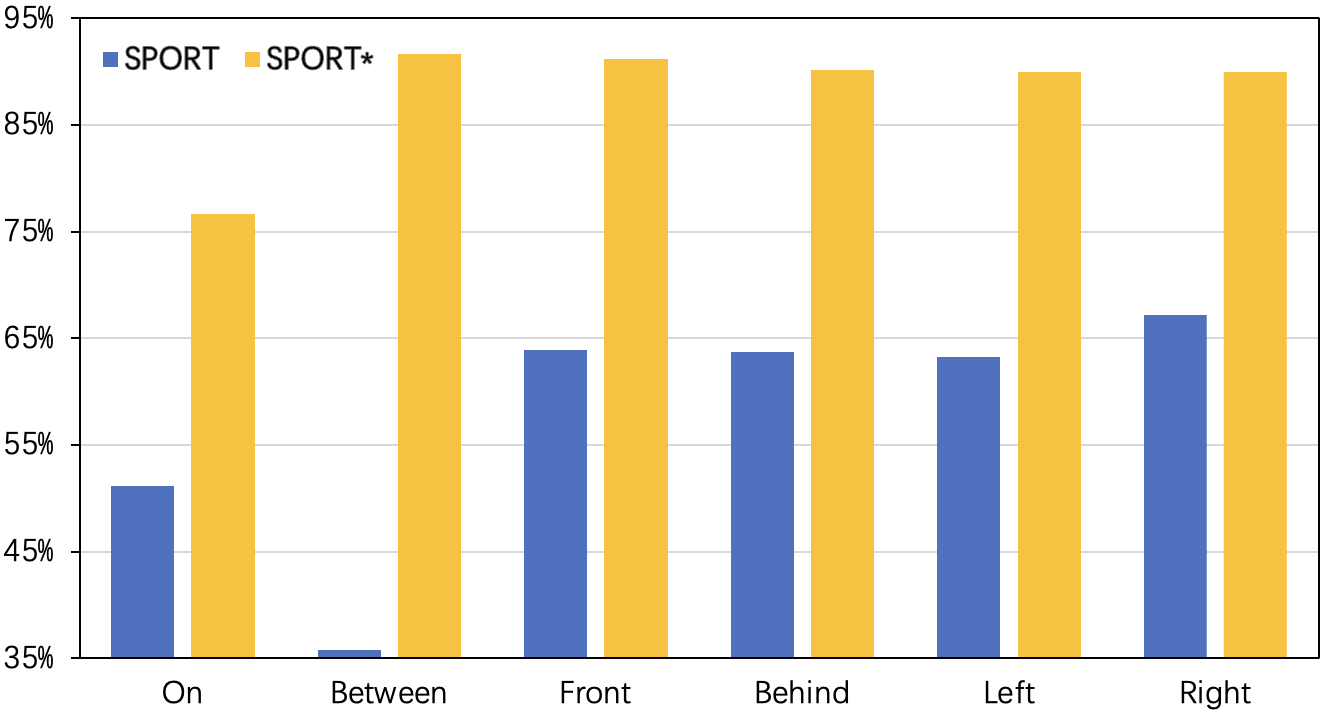}
  \caption{Performance of six specific scenarios in simulation.}
  \label{Fig:6_relation}
\end{figure}

Six scenarios (corresponding to six spatial relationships) are conducted for evaluation. The results are listed in Table \ref{table:1}. Whether a testing sample is classified as correct depends on three aspects: the objects are recognized, the generated object poses satisfy the instruction, and the rearrangement is physically-realistic. ``Pose Accuracy'' refers to the accuracy of the first two aspects, ``Physical Realism'' focuses on the last one, and ``Overall Success'' is the overall success rate considering all three aspects. Moreover, the ``Pose Accuracy'' for each scenario are shown in Fig. \ref{Fig:6_relation}.

It can be seen that SPORT achieves an overall success rate of $46.19\%$. The result is acceptable due to the challenging experimental setting: the testing and training data are from different datasets (\textit{the 77 testing objects are unseen to the model}), and the final states of objects should be physically valid. However, we want to explore more, especially given the observation that Physical Realism achieve higher accuracy than Pose Accuracy, which is quite unusual.

We notice that LISA fails a lot on simulation-rendered images. The main reason is the substantial domain differences between the simulation images and the real-world images used in LISA's training set. On rendered images, the completeness of the object segmentation masks produced by LISA is lacking, and thus the quality of the resultant 3D point clouds is often poor. However, LISA can indeed localize target objects even requiring complex reasoning in real-world scenarios. We have conducted real-world experiments, please refer to subsection \ref{sec:real-world} for the details.

According to this observation, we want to know the performance if the required objects can be successfully obtained in simulation (because LISA can achieve this on real-world images), by using object masks directly from simulation data. The results are listed in the second row in Table \ref{table:1}, abbreviated as SPORT*. It achieves convincing performance, $87.8\%$ on Pose Accuracy and $69.49\%$ on Overall Success.

Finally, our method demonstrates a certain degree of~effectiveness in generating physically-realistic poses. This is a challenging task, requiring the generative model to directly estimate stable, collision-free and physically feasible poses. SPORT* achieves $76.40\%$ on Physical Realism, assessed by using the Pybullet simulator. 

\textbf{Ablation study}

The baseline of ablation studies is SPORT*.

\textit{Estimating Poses of Reference and Irrelevant Objects}. In our approach, the poses of the reference and irrelevant objects are fixed when training the diffusion model. An exper-iment is conducted to explore the impact of this design. The performance of not fixing these poses is listed in the third row in Table \ref{table:1}. The success rate decreases from $69.49\%$ to $65.59\%$. Such a design helps the model identify the reference points and possible collisions, focusing on the learning of replacing the target object according to the command. 

\textit{Making the instruction encoder trainable}. We train BERT in the goal pose estimator to study its effect on the performance. As listed in the fourth row in Table \ref{table:1}, the performance decreases significantly. This is really interesting since end-to-end tuning is quite standard. The probable reason is that our collected data is insufficient to train all network modules, while pre-trained BERT is already good enough for text understanding. Similar observations can be found in \cite{Detic}. 

\textit{Training on different scales of data}. The effect of different scales of training data on model performance is studied here. Three models, trained with 10\%, 25\%, and 50\% of the entire dataset, are compared. The results in Table \ref{table:1} indicate a clear trend: as the volume of training data increases, the model's performance correspondingly improves. We do not further study the scaling law due to the extensive resources and time required. We leave it for future work.

\subsection{Real-world Experiments}

\begin{figure}
  \centering
  \includegraphics[width=0.45\textwidth]{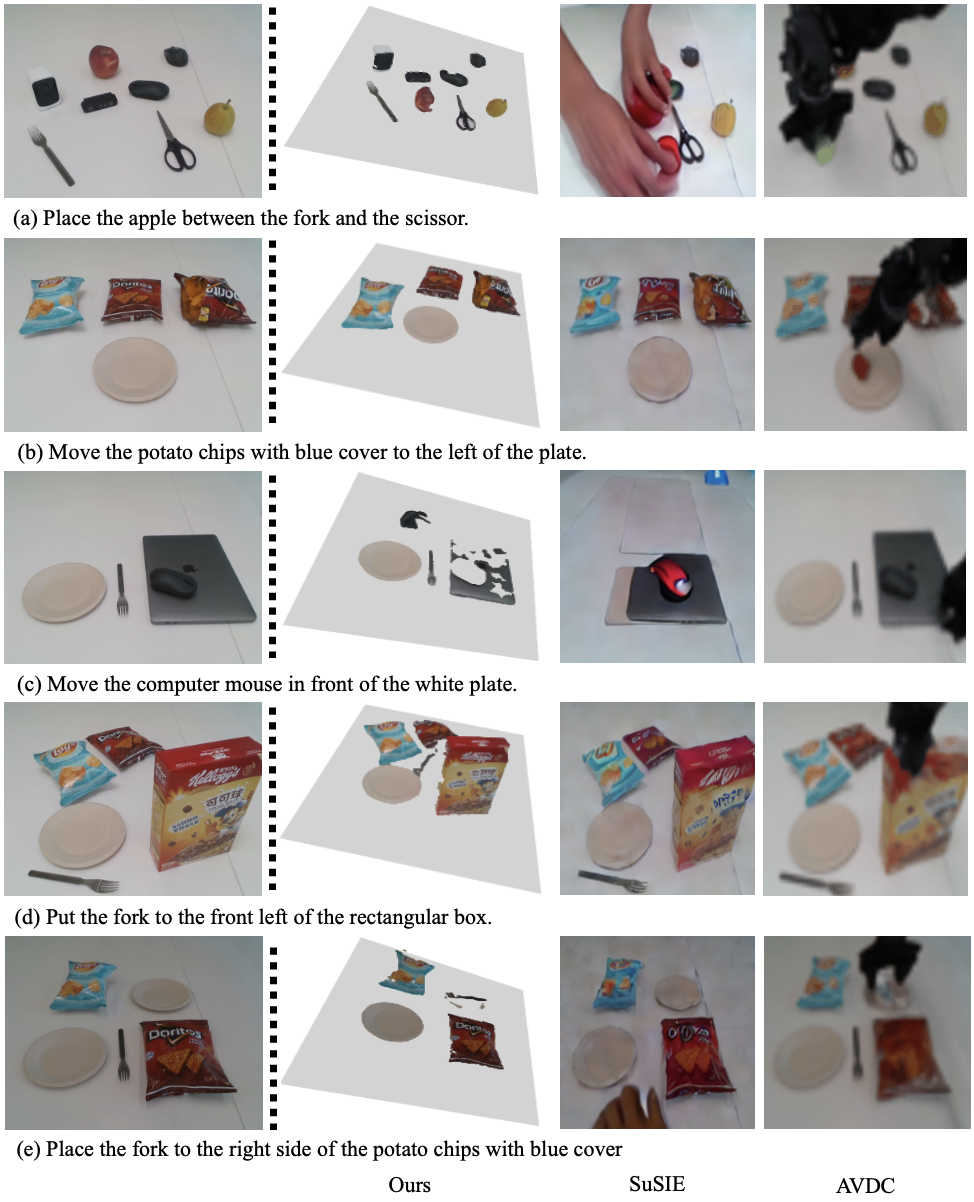}
  \caption{Results for the five tasks on the real-world planar scene. The first column shows initial scenes captured with a RGB-D camera. Our model is trained in simulation and tested directly on these real-world cases.}
  \label{fig:real-world-exp}
\end{figure}

\label{sec:real-world}

\textbf{Setup} We collect 10 real-world scenes, captured by an Intel RealSense D435i RGB-D camera. Accomplishing these real-world tasks requires complex spatial reasoning involving multiple, yet similar, objects. Moreover, the last five scenes include objects placed on a 3D shelf, which is very different from the training scenes and thus more challenging.

\textbf{Competitors}

\textit{SuSIE} \cite{susie} and \textit{AVDC} \cite{AVDC} both leverage diffusion models (\emph{e.g.}, \cite{InstructPix2Pix}) to synthesize the imagined execution process of placement. The former generates subgoal images in an ``edit'' manner while the latter synthesizes a video directly.

\begin{figure}
  \centering
  \includegraphics[width=0.48\textwidth]{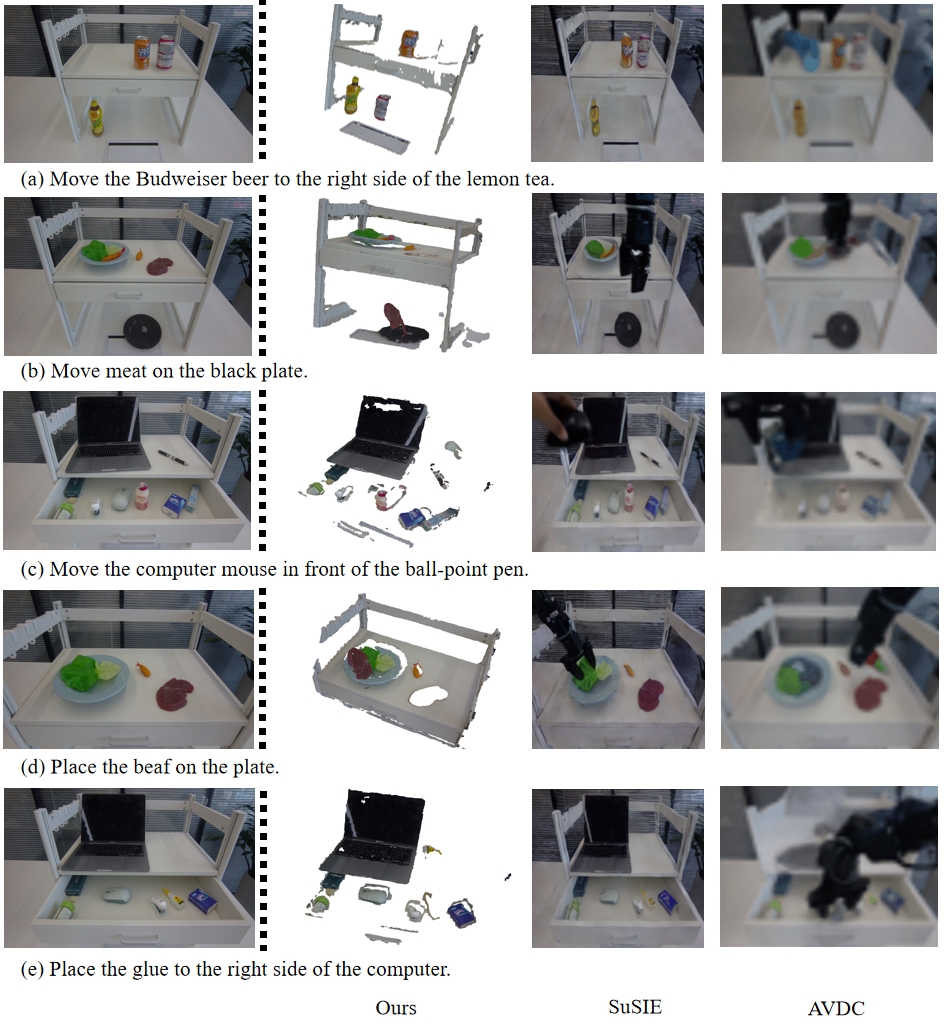}
  \caption{Results for the five tasks on the real-world 3D shelf scene. This kind of task that requires placing objects within 3D structures is more challenging because the model has not encountered it during training.}
  \label{fig:real-world-exp2}
\end{figure}

\textbf{Performance comparison}

Fig. \ref{fig:real-world-exp} and Fig. \ref{fig:real-world-exp2} illustrate the results. SPORT performs significantly better than the competitors. It can ``imagine'' the object positions strictly following the command, performing the rearrangement directly in 3D space. For example, SPORT can identify Budweiser beer from canned drinks and predict its accurate pose, even if the placement needs to be performed within a 3D structure (the first row in Fig. \ref{fig:real-world-exp2}). 

In comparison, AVDC produces fuzzy and less-realistic videos, and often fails to generate correct goal images. As for SuSIE, actually the results cannnot be fully reproduced since we do not have the authors' robot setup. Nevertheless, we use the source code to produce goal images by repeating the image generation several times. Some issues can still be observed to some extent, \emph{e.g.}, hallucination and loss of details. Moreover, both of these methods would hallucinate a robot or a human arm, originated from training data, which we believe may affect generalization.

Although the main contribution of this work is the goal pose prediction, we also demonstrate that a robot can indeed pick and place objects based on these goal poses. See our accompanying videos for demos. In the demo, Groundingdino \cite{Groundingdino}, GraspNet \cite{graspnet} and MoveIt \cite{moveit} are used for detection, grasping and motion planning, respectively.

\section{Conclusions}

In this paper, we demonstrate the potential of achieving general-purpose object rearrangement with precise ``something something'' instructions. Specifically, given an instruction in natural language format, an LLM is used to identify the objects to be moved and to be the reference. Given RGB-D images, we utilize a pre-trained LLM-enhanced image segmentation model to obatin required objects and then point clouds. Based on these results, a diffusion-based 3D pose estimation model can follow precise low-level instructions to achieve physically-realistic position predictions. By establishing a GPT-assisted pipeline from a 3D perspective, a high-quality dataset for the placement task is generated. The results from both simulation and real-world experiments demonstrate the effectiveness of our approach. The model trained with simulation data can seamlessly transfer to real-world scenarios, achieving promising performance.

This project is a work still in progress, and several directions can be explored: (1) \textit{More realistic data}. More realistic object models sampled from diverse environments and scenarios may probably benefit the model learning. Real-to-sim methods are worth trying. (2) \textit{Further improve physical realism}. More strategies could be developed, \emph{e.g.}, encoding gravitational field, with which the model can simulate and predict a greater variety of real-world physical laws. This may be a big step to the ``world model''. (3) \textit{More powerful models}. More recent diffusion models with high-capacity could be utilized to better estimate the object poses.

\bibliographystyle{IEEEtran}
\bibliography{IEEEabrv, ref}

\end{document}